\def\year{2017}\relax
\documentclass[letterpaper]{article}
\usepackage{aaai17}
\usepackage{times}
\usepackage{helvet}
\usepackage{courier}
\usepackage{graphicx}
\usepackage{subfigure}
\usepackage{CJKutf8}
\usepackage{array}
\begin{document}
%
\title{Structural Correspondence Learning for Cross-lingual Sentiment Classification with One-to-many Mappings}
\author{Nana Li\\School of Computer Science and Engineering \\Hebei University of Technology\\ Tianjin 300401, China\\lin@binghamton.edu
\And
Shuangfei Zhai, Zhongfei Zhang\\Computer Science Department\\Binghamton University \\ Binghamton, NY 13902,USA\\szhai2@binghamton.edu, zhongfei@cs.binghamton.edu\\
\And 
 Boying Liu\\ School of Electrical Engineering\\Hebei University of Technology\\Tianjin 300401, China\\lby@hebut.edu.cn\\
}
\maketitle
\begin{abstract}
Structural correspondence learning (SCL) is an effective method for cross-lingual sentiment classification. This approach uses unlabeled documents along with a word translation oracle to automatically induce task specific, cross-lingual correspondences. It transfers knowledge through identifying important features, i.e., pivot features. For simplicity, however, it assumes that the word translation oracle maps each pivot feature in source language to exactly only one word in target language. This one-to-one mapping between words in different languages is too strict. Also the context is not considered at all. In this paper, we propose a cross-lingual SCL based on distributed representation of words; it can learn meaningful one-to-many mappings for pivot words using large amounts of monolingual data and a small dictionary. We conduct experiments on NLP\&CC 2013 cross-lingual sentiment analysis dataset, employing English as source language, and Chinese as target language. Our method does not rely on the parallel corpora and the experimental results show that our approach is more competitive than the state-of-the-art methods in cross-lingual sentiment classification.
\end{abstract}

\section{Introduction}
Sentiment classification is the task to predict the sentiment polarity of a given document such as a product review or commentary essay. Its goal is to develop automated approaches that can classify sentiment polarity in text as positive, neutral, or negative. To obtain a satisfactory classification performance, most methods require lots of labeled data, which can be costly in terms of both time and human efforts. While there have been lots of resources available in English, including labeled corpora and sentiment lexicons, for other languages, such resources are often insufficient. Thus, it is expected to make use of the knowledge learned from those resource-rich languages to perform sentiment classification in other languages, which can substantially reduce human efforts. This problem is called cross-lingual sentiment classification (CLSC), which we address in this paper.

CLSC uses annotated sentiment corpora in one language as the training data, to predict the sentiment polarity of the data in another language. Domain adaptation focuses on solving this problem by transferring knowledge from the typically sufficient training samples in a resource-rich language to target, resource-scarce language. As a result, domain adaptation has been proposed to address this problem  (\cite{blitzer2006domain}; \cite{banea2008multilingual}; \cite{prettenhofer2011cross}). There has been a lot of work in domain adaptation, and one effective method for CLSC is based on structural correspondence learning (SCL), named as cross-lingual structural correspondence learning (CL-SCL) proposed by Prettenhofer and Stein \cite{prettenhofer2010cross}. Its key idea is to identify a low-dimensional representation that captures the correspondence between features from both domains by modeling their correlations with some special pivot features. From these correspondences a cross-lingual representation is created that enables the transfer of classification knowledge from the source to the target language. This approach is a good fit for CLSC as it transfers knowledge through identifying important features. However, for simplicity, CL-SCL assumes that the word translation oracle maps each pivot word in source language to exactly only one word in target language. As we all know, machine translation performs simple substitution of words in one language for words in another, but that alone usually cannot produce a good translation. Furthermore, this one-to-one translation between words in different languages is too strict.
\par In 2013, Mikolov et al.\cite{mikolov2013exploiting} proposed a method for exploiting similarities among languages. It used the distributed representation of words and learned a linear mapping between vector spaces that represent the corresponding languages, respectively. It translated word and phrase entries by learning language structures based on large monolingual data and mappings between languages from small amount of bilingual data. Despite its simplicity, the results showed that their method was surprisingly effective, especially for the translation between languages that are substantially different (such as English to Chinese). In this paper we introduce the distributed representation of words into the CL-SCL, and propose a novel structural correspondence learning method with one-to-many mappings (SCL-OM). This method aims at building up the one-to-many mappings between the pivot features in source language and those in the target language. Evaluations on NLP\&CC 2013 datasets show that our algorithm outperforms the state-of-the-art methods.

\section{Related Work}
\subsection{Sentiment Classification}
Sentiment classification is usually formulated as a two-class classification problem, positive and negative. Training and testing data are normally product reviews. It has emerged and become a very active research area since the year 2000 \cite{liu2012sentiment}. In general, sentiment classification has been investigated mainly at three levels: document level, sentence level, and aspect level. In this paper, we only focus on document level. Document level sentiment classification aims to classify an opinion document as expressing a positive or negative opinion. The approaches are generally based on two kinds of resources: sentiment lexicons and corpora. Lexicon-based approaches predict the sentiment polarities by creating and using sentiment lexicons, while corpora-based approaches generally treat the sentiment classification problem as a machine learning task. Most of the existing approaches focus on extracting various features from text and then applying supervised learning techniques to learn classifiers \cite{hu2004mining};\cite{mullen2004sentiment}. Pang \cite{pang2002thumbs} was the first to take supervised learning to classify movie reviews using unigrams as features in classification with standard machine learning techniques (Naive Bayes, maximum entropy classification, and support vector machines). In the subsequent research, many more features and learning algorithms were developed by a number of researchers. Like other supervised machine learning applications, the key factor for sentiment classification is a set of effective features. There are also some unsupervised methods. Turney et al. \cite{turney2002thumbs} were the first to apply an unsupervised learning technique based on the mutual information between document phrases and predict the sentiment orientation by the average scores of the phrases given in the document. Zhai et al. \cite{zhai2015semisupervised} proposed an autoencoder based semisupervised learning method to learn representations with both labeled and unlabeled data. However, supervised approaches are the mainstream methods for sentiment classification. Moreover, most of the existing supervised and semi-supervised approaches typically require high-quality labeled data to train classifiers with a good accuracy. 
\subsection{Cross-lingual Sentiment Classification}
The traditional CLSC approaches employ machine translation systems to bridge the gap between the source language and target language. Wan et al. \cite{wan2009co} proposed a co-training approach to address this problem. The labeled English reviews and unlabeled Chinese reviews were translated into labeled Chinese reviews and unlabeled English reviews separately. Each review thus had the two views. support vector machines (SVM) was then applied to learn two classifiers. Finally, the two classifiers were combined into a single classifier. Pan et al. \cite{pan2011cross} designed a bi-view non-negative matrix tri-factorization model using machine translation. It learned previously unseen sentiment words from the large parallel dataset. Li et al. \cite{li2011semi} studied semi-supervised learning for imbalanced sentiment classification by using a dynamic co-training approach. Gui et al. \cite{gui2013mixed} compared several of these approaches, and then they incorporated class noise detection into transductive transfer learning to reduce negative transfers in the process of transfer learning \cite{gui2014cross}. However, machine translation is far from perfect. The translated text can potentially mislead the classifier. Consequently, many researchers use domain adaptation to solve this problem. Various domain adaptation techniques are explored \cite{blitzer2007biographies}; \cite{banea2008multilingual}; \cite{prettenhofer2011cross}; \cite{meng2012cross}. Prettenhofer and Stein \cite{prettenhofer2010cross} proposed a representative domain adaptation approach CL-SCL which was effective for cross-lingual sentiment classification. They found a set of pivot features shared by both source language and target language, and then learned the correlations between pivot and non-pivot features and generated a projection matrix to build a bridge between the two languages.  Meng et al. \cite{meng2012cross} proposed a cross-lingual mixture model (CLMM) to leverage unlabeled bilingual parallel data. Zhang et al. \cite{zhang2015semi} proposed a semi-supervised learning approach with an adjusted method to train an initial classifier to predict the labels for target instances and then to obtain a new label space and a large-scale, labeled target language dataset. It selected the confident instances and trained a new classifier. This method needs to learn three classifiers and is very time-consuming. 

With the development of deep learning, shared deep representations are employed for CLSC. Some researchers apply deep learning techniques to learn bilingual representations \cite{zhou2015learning};\cite{mogadalabilingual};\cite{zhoucross}. Paired sentences from parallel corpora are used to learn word embeddings across languages, eliminating the need of machine learning. Zhou et al. \cite{zhou2015learning} proposed an approach to learning bilingual sentiment word embeddings (BSWE) for English-Chinese CLSC, and it incorporated sentiment information of text into bilingual embeddings. Zhou et al. \cite{zhoucross} proposed a cross-lingual representation learning model which simultaneously learned both the word and document representations in both languages. However, high-quality bilingual embeddings rely on the large-scale task-related parallel corpora, which are also a scarce resource.

\begin{figure*}[t]
\centering
\includegraphics[scale=0.56]{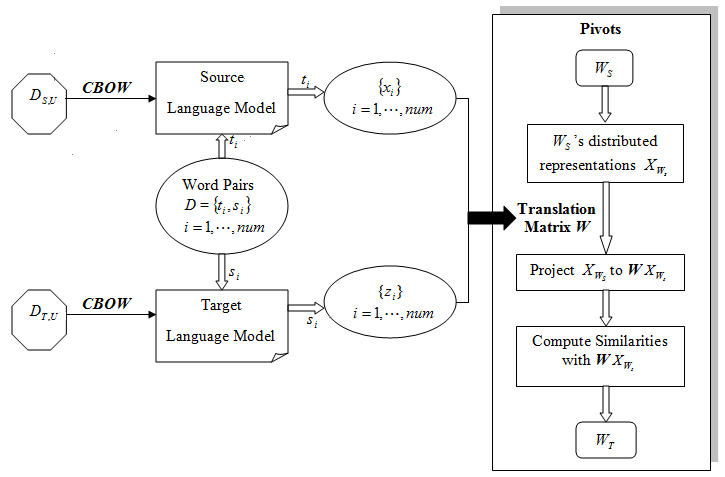}
\caption{The Framework of Generating Pivot Features.}
\label{fig:model}
\end{figure*}
\section{CL-SCL with One-to-many Mappings}
\subsection{Problem Definition}
We have a set of labeled training documents $D_S = \{(x_i, y_i)\}_{i=1}^n$ written in source language $S$; $X_S$ is the source language feature space, $X_T$ is the target language feature space, and $Y$ is the set of class labels. Let $X=X_s\bigcup X_T$ denote the feature space. For simplicity, and without loss of generality, we consider the binary classification problem, i.e., $Y=\{ +1, -1\}$. In addition, $V_S$ denotes the source vocabulary, $V_T$ denotes the target vocabulary, and $V=V_S\bigcup V_T$. Besides the labeled training documents $D_S$, we have unlabeled documents $D_{S,U}$ and $D_{T,U}$ from the source language $S$ and the target language $T$, respectively. Let $D_U$ denote $D_{S,U}\bigcup D_{T,U}$. The goal is to create a classifier for documents written in a different language $T$, which can predict the labels of new, previously unseen reviews in $T$.

\subsection{Pivot Sets Based on CL-SCL with One-to-many Mappings}

%
%

Our approach is based on CL-SCL, which is an approach to cross-lingual text classification that builds on structural correspondence learning. In order to induce task-specific, cross-lingual word correspondences, this approach used unlabeled documents, along with a word translation oracle. The advantage of this method is resource efficiency because parallel corpora are not recquired. The first step of CL-SCL is to define a set of pivot features on the unlabeled data from both languages. Then these pivot features are used to learn a mapping $ \theta $ from the original feature spaces of both languages to a shared, low-dimensional feature space. Once such a mapping is found, the cross-lingual sentiment classification problem is reduced to a standard classification problem in the cross-lingual space. 

Pivot features are selected to induce correspondences among the words from both languages, and they play a very important role in CL-SCL. A pivot is a pair of words, $\{w_S, w_T\}$, separately from the source language $S$ and the target language $T$, which possess the similar semantics. $w_T$ is $w_S$’s translation in the target vocabulary $V_T$ by querying the translation oracle. However, for simplicity, CL-SCL assumes that the word translation oracle maps each pivot in source language into exactly only one word in target language. This one-to-one mapping between words in different languages is too strict. In addition it does not consider the context either when translating the pivot features. In this step we propose a cross-lingual SCL based on distributed representation of words; it learns meaningful one-to-many translations for words using large amounts of monolingual data and a small dictionary. The framework of our approach to generate pivot features is demonstrated in Figure \ref{fig:model}.

First, select pivots $w_S$ in the source language according to the mutual information with respect to the class labels in the labeled training documents in source language. CL-SCL used a word translation oracle (e.g., a domain expert) to map words in the source vocabulary $V_S$ to their corresponding translations in the target vocabulary $V_T$. Unlike CL-SCL, it is more reasonable to build one-to-many mappings between words in the two languages. Thus, instead of using translator or domain expert, we build a one-to-many mapping by learning from bilingual word pairs using distributed representation of words. We incorporate the method Word2Vec recently proposed in \cite{mikolov2013exploiting} into CL-SCL for the translation. We use Word2Vec's CBOW model to learn the representations of languages. This model learns word representations using the neural network architecture that aims to predict the neighbors of a word. The process is as follows:
\begin{itemize}
\item Step 1, Build monolingual models of languages (i.e., distributed representation of words in source language and target language) using large amounts of documents $D_{S,U}$ and $D_{T,U}$. Suppose that there are a set of word pairs $D=\{s_i,t_i\} (i=1,2,\cdots,num)$, $s_i\in V_S$, $t_i\in V_T$, where $t_i$ is the translation of $s_i$. We obtain their associated vector representations $\{x_i,z_i\}, $ where $x_i$ is the distributed representation of word $s_i$ in the source language $S$, and $z_i$ is the vector representation of $t_i$ in the target language $T$.
\item Step 2, Use  $\{x_i,z_i\}$ to learn a linear projection between the languages. The goal is to find a translation matrix $W$ such that $Wx_i$ approximates $z_i$. We learn $W$ by the following optimization problem:
\begin{equation}
\min\limits_{W}  \sum_{i=1}^{num} \Vert{Wx_i-z_i}\Vert ^2_2
\end{equation}

\item Step 3, Translate $w_S$  into $w_T$ by projecting their vector representations from the source language space to the target language space. Suppose that the vector representation of word $a$ $(a\in w_s)$ is $x_{as}$. We map it to the target language space by computing $b =Wx_{as}$. Then we obtain a set $\Psi$ of $p_n$ words that are the top $p_n$ closest to $b$ in the target language space, using cosine similarity as the distance metric. For simplicity, we set $p_n=3$. $\Psi=\{(word_1,d_1),(word_2,d_2),(word_3, d_3)\}$, where $d_i$ denotes the cosine distance between $b$ and $word_i$, and they can be automatically obtained from Word2Vec. We define a  threshold $\phi$. If $d_1-d_2$ and $d_2-d_3$ are both smaller than $\phi$, we take $\{word_1,word_2,word_3\}$ as the translation of $a$; if $d_1-d_2$ is smaller than $\phi$ but $d_2-d_3$ is larger than $\phi$, we take $\{word_1,word_2\}$ as the translation; otherwise, we take $\{word_1\}$ as the translation of $a$. For example, we consider the pivot words ``excellent'' and ``recommend'' as follows: 
$\Psi_{excellent}=$
\begin{CJK*}{UTF8}{gbsn}
{\{(棒/excellent,0.628),(太好了/very good, 0.613), (出色/outstanding, 0.603)\}}
\end{CJK*}; and $\Psi_{recommend}=$\begin{CJK*}{UTF8}{gbsn} {\{(推荐/recommend, 0.835), (建议/advise, 0.695), (购买/buy, 0.581)\}}\end{CJK*}. The English word after ``/'' is the translation from chinese to english for non-chinese readers.
Supposing that $\phi$ is 0.05, the translations of ``excellent'' and ``recommend'' respectively are :\begin{CJK*}{UTF8}{gbsn}
{\{棒,太好了,出色\}}
\end{CJK*} and \begin{CJK*}{UTF8}{gbsn}
{\{推荐\}}.
\end{CJK*}

\end{itemize}

Finally, we eliminate the candidate pivots $\{w_S,w_T\}$ where the document frequency of $w_S$ or $w_T$ is smaller than a threshold $\delta$.

\subsection{Framework of the Proposed Method}
First, as described above, we generate pivot features $P(\vert P \vert = m)$ which are pairs of words $\{w_S,w_T\}$, where $w_S$ is the pivot word in source language and $w_T$ is the pivot word in target language. From $w_S$ to $w_T$, it is a one-to-many mapping by learning from bilingual word pairs using distributed representation of words. The details are described in the above Section.
Second, similar to CL-SCL, we build the connection from unlabeled documents in both source and target languages and obtain the low-dimensional hypothesis space $\theta$. For each pivot $p_l \in P$ , a linear classifier is trained to model the correlations between the pivot $\{w_S,w_T\}$ and all other words $w \notin \{w_S,w_T\}$. Each linear classifier is characterized by the parameter vector $w_l$. Thus, a $\vert V \vert \times m$ dimensional parameter matrix $W$ can be obtained, $W=[w_1w_2\cdots w_m]$. Correlations across pivots are identified by computing the singular value decomposition of $W$ to find a low dimensional representation.
\begin{equation}
U \Sigma V^T=SVD(W)
\end{equation}

Choosing the columns of $U$ associated with the largest singular values yields those substructures that capture most of the correlation in $W$. Define $\theta$ as those columns of $U$ that are associated with the $k$ largest singular values:
\begin{equation}
\theta=U_{[1:k,1:\vert V \vert ]}^\mathrm{T}
\end{equation}

Apply the projection $\theta$ to each input instance $x$. The vector $v$ that minimizes the regularized training error for $D_S$ in the projected space is defined as follows:
\begin{equation}
v^* = \mathop{argmin} \limits_{v \in R^k}\sum \limits_{(x,y) \in D_S}L(y,v^T\theta x) + \frac{\lambda}{2} \Vert v \Vert ^2
\end{equation}

The final classifier $f_{ST}(x)$ is defined as follows:
\begin{equation}
f_{ST}(x)=sign(v^{*\mathrm{T}}\theta x)
\end{equation}

\section{Experiments}
In this section, we evaluate the effectiveness and efficiency of the algorithm SCL-OM proposed in this paper. We use English as source language and Chinese as target language for the task of cross-lingual sentiment classification. 
\subsection{Dataset and Preprocessing}
We evaluate the proposed approach on an open cross-lingual sentiment analysis task in NLP\&CC 2013. The dataset includes product reviews of three product categories from Amazon (Books, DVD, and Music). Each category contains 4,000 labeled English reviews as the training data, 4,000 Chinese reviews as the test data, and over ten thousands of Chinese product reviews without label. Furthermore, since training the monolingual language model needs a large amount of text data, we use the unlabeled English reviews from \cite{prettenhofer2011cross} to learn the representations of English words. See Table \ref{tb:data_sets} for details.
\begin{table}[ht]
\normalsize
\centering
\begin{tabular}{|c|c|c|c|}
\hline
Data & Books & DVD & Music 
\\
\hline
labeled English reviews & 4000 & 4000 & 4000
\\
\hline
unlabeled English reviews & 49999 & 30000 & 25220
\\
\hline
unlabeled Chinese reviews & 47071 & 17814 & 29677
\\
\hline
Chinese test reviews & 4000 & 4000 & 4000
\\
\hline
\end{tabular}
\caption{The Data sets.}
\label{tb:data_sets}
\end{table}

Each English or Chinese review includes \emph {summary}, \emph {text} and \emph {category}; we extract the content of \emph {summary} and \emph {text} and combine them as one review document $d$ , which is expressed as a feature vector $x$ using unigram bag-of-words model. In addition, we only select those words as the features with the frequency $fre_w$ larger than 5. We summarize the vocabulary size of the datasets in Table \ref{tb:vocab_size}.
\begin{table}[ht]
\normalsize
\centering
\begin{tabular}{|c|c|c|c|}
\hline
Vocabulary size & Books & DVD & Music 
\\
\hline
$\vert V_S \vert$ & 35966 & 24588 & 17439
\\
\hline
$\vert V_T \vert$ & 16998 & 7460 & 11248
\\
\hline
\end{tabular}
\caption{Vocabulary size.}
\label{tb:vocab_size}
\end{table}

\begin{figure*}[t]
\centering
\includegraphics[scale=0.6]{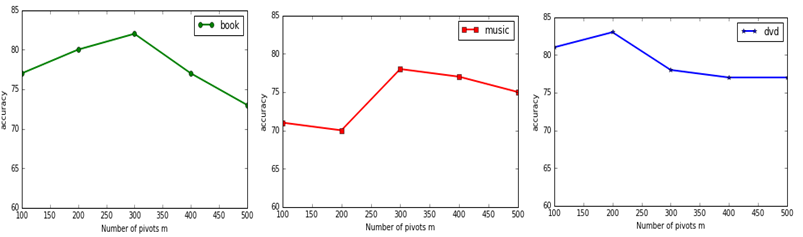}
\caption{Influence of the number of pivots $m$ on the performance of SCL-OM.}
\label{fig:m}
\end{figure*}
\begin{figure*}[t]
\centering
\includegraphics[scale=0.6]{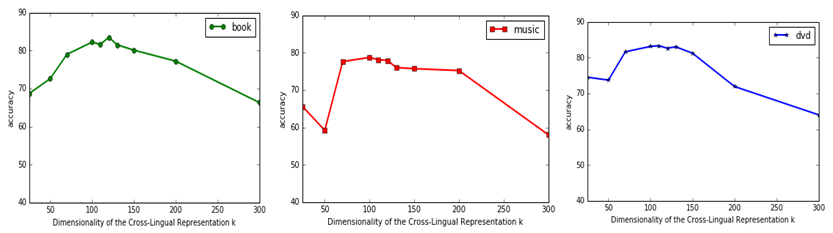}
\caption{Influence of the dimensionality of the cross-lingual representation $k$ on the performance of SCL-OM.}
\label{fig:k}
\end{figure*}
First, the monolingual word vectors are trained using CBOW model with negative sampling of window size 5. To generate a bilingual dictionary between languages, we use the most frequent 500 words from the monolingual source datasets, and translate these words using online Google Translate. In addition, in our experiments, the Chinese word segmentation tool is Jieba, and the monolingual sentiment classifier is SVM.

\subsection{Methods}
\begin{itemize}
\item Train\_CHN: The labeled English reviews are translated to Chinese by Google Translate with correspondence labels. A Chinese SVM classifier is learned with the translated reviews. The Chinese testing dataset is used for test.
\item Train\_ENG: using labeled English reviews as training data, an English SVM classifier is learned. Then the Chinese test reviews are translated to English by Google Translate. The translated testing dataset is used for test.
\item Basic co-training (Co-Train): The co-training method proposed in \cite{pan2011cross} is implemented. It is a bidirectional transfer learning.
\item 	CL-SCL: This method is proposed by Prettenhofer et al. Google Translate is used to do the translation which only returns one word for each pivot feature. Also we fix the same parameter values as those in our method $m=300$, $\varphi=30$, $k=120$.
\item 	Best result in NLP\&CC 2013 : This is the best result reported in NLP\&CC 2013. Unfortunately, the specification of the method is not available.
\item NTD (Best): Gui et al. \cite{gui2013mixed} proposed a mixed CLSC model by combining co-training and transfer learning strategies. They \cite{gui2015improving} further improved the accuracy by removing the noise from the transferred samples to avoid negative transfers (NTD).
\item BSWE (Best): Zhou et al. \cite{zhou2015learning} proposed a method that learned bilingual sentiment word embedding (BSWE) for English-Chinese CLSC. The proposed BSWE incorporated sentiment information of text into bilingual embedding.
\end{itemize}

The baseline methods described above are categorized into two classes: the first four are preliminary methods; the last three are state-of-the-art models for CLSC. All the methods use SVM as basic classifier and use unigram+bigram features to train the basic classifiers except that CL-SCL and our approach use unigram features. 
\begin{table}[ht]
\normalsize
\centering
\begin{tabular}{|c|c|c|c|c|}
\hline
Method & Book & DVD & Music & Average
\\
\hline
Train\_CHN & 0.544 & 0.583 & 0.607 & 0.671
\\
\hline
Train\_ENG & 0.776 & 0.759 & 0.738 & 0.728
\\
\hline
Co-Train(Best) & 0.796 & 0.804 & 0.783 & 0.794
\\
\hline
CL-SCL & 0.780 & 0.772 & 0.739 & 0.763
\\
\hline
NLP\&CC 2013 (Best) & 0.785 & 0.777 & 0.751 & 0.771
\\
\hline
NTD (Best) & 0.805 & 0.822 & \textbf{0.797} & 0.808
\\
\hline
BSWE (Best) & 0.810 & 0.816 & 0.794 & 0.807
\\
\hline
SCL-OM & \textbf{0.829} & 0.826 & 0.787 & 0.814
\\
\hline
SCL-OM(Best) & \textbf{0.829} & \textbf{0.833} & 0.787 & \textbf{0.816}
\\
\hline
\end{tabular}
\caption{Performance comparisons on the NLP\&CC 2013 CLSC data set.}
\label{tb:comparisons}
\end{table}
\begin{table*}[ht]
\normalsize
\centering
\begin{tabular}{|p{0.1\textwidth}|p{0.35\textwidth}|p{0.09\textwidth}|p{0.35\textwidth}|}
\hline 
excellent & \begin{CJK*}{UTF8}{gbsn}{棒/excellent；太好了/very good；

出色/outstanding}\end{CJK*} & amazing &  \begin{CJK*}{UTF8}{gbsn}{吃惊/surprise；惊讶/amazing；

不敢相信/unbelievable}\end{CJK*}
\\
\hline
great & \begin{CJK*}{UTF8}{gbsn}{很棒/great; 

超值/ good value }\end{CJK*} & awesome &  \begin{CJK*}{UTF8}{gbsn}{太棒了/ awesome；太好了/very good； 

超棒/ terrific}\end{CJK*}
\\
\hline
good &   \begin{CJK*}{UTF8}{gbsn}{好/good；不错/not bad}\end{CJK*} & save & \begin{CJK*}{UTF8}{gbsn}{节省/save}\end{CJK*} 
\\
\hline
waste &  \begin{CJK*}{UTF8}{gbsn}{浪费/waste}\end{CJK*} &  terrible & \begin{CJK*}{UTF8}{gbsn}{糟糕/ terrible；不好/not good；差/bad}\end{CJK*}
\\
\hline
disappointed &  \begin{CJK*}{UTF8}{gbsn}{失望/disappoint}\end{CJK*} & stupid &  \begin{CJK*}{UTF8}{gbsn}{傻/stupid；愚蠢/fool}\end{CJK*}
\\
\hline
boring &  \begin{CJK*}{UTF8}{gbsn}{无聊/boring；乏味/tedious；空洞/inanity}\end{CJK*} &  horrible & \begin{CJK*}{UTF8}{gbsn}{太差/too bad；太烂/horrible}\end{CJK*}
\\
\hline
worst & \begin{CJK*}{UTF8}{gbsn}{最差/worst；最烂/worst}\end{CJK*} & awful & \begin{CJK*}{UTF8}{gbsn}{差劲/awful；糟糕/terrible}\end{CJK*}
\\
\hline
bad & \begin{CJK*}{UTF8}{gbsn}{差/bad；差劲/awful；不好/not good}\end{CJK*} & recommend &  \begin{CJK*}{UTF8}{gbsn}{推荐/recommend}\end{CJK*}
\\
\hline
fact & \begin{CJK*}{UTF8}{gbsn}{事实/fact}\end{CJK*} & no & \begin{CJK*}{UTF8}{gbsn}{没有/no；不/not}\end{CJK*}
\\
\hline
wonderful & \begin{CJK*}{UTF8}{gbsn}{精彩/wonderful；棒/excellent} \end{CJK*} &  money & \begin{CJK*}{UTF8}{gbsn}{钱/money}\end{CJK*}
\\
\hline
highly & \begin{CJK*}{UTF8}{gbsn}{极其/highly；非常/very；十分/very} \end{CJK*} & fun & \begin{CJK*}{UTF8}{gbsn}{好玩/fun；有趣/interesting；搞笑/funny} \end{CJK*}
\\
\hline
journey & \begin{CJK*}{UTF8}{gbsn}{旅行/travel；旅程/journey；旅途/journey} \end{CJK*} & beautiful & \begin{CJK*}{UTF8}{gbsn}{漂亮/beautiful；可爱/lovely；美/beautiful} \end{CJK*}
\\
\hline
ridiculous & \begin{CJK*}{UTF8}{gbsn}{可笑/ridiculous；扯淡/nonsense} \end{CJK*} & life & \begin{CJK*}{UTF8}{gbsn}{生活/life；人生/life} \end{CJK*}
\\
\hline
useless & \begin{CJK*}{UTF8}{gbsn}{无用/useless；没用/useless} \end{CJK*} & like & \begin{CJK*}{UTF8}{gbsn}{喜欢/like}\end{CJK*}
\\
\hline
\end{tabular}
\caption{Visualization of some learned pivots.}
\label{tb:pivots}
\end{table*}
\subsection{Performance Results}
Recall that SCL-OM has six parameters as input: the number of pivots $m$, the dimensionality of the cross-lingual representation $k$, the minimum support $\delta$ of a pivot, word similarity distance threshold $\phi$, the dimensionality of English word vectors and the dimensionality of Chinese word vectors. We use fixed values of $m=300$, $k = 120$, $\delta = 30$, and $\phi = 0.1$. As for the dimensionality of word vectors, Mikolov et al. \cite{mikolov2013exploiting} showed that the dimensionality of the vectors trained in the source language should be several times (around 2 to 4 times) larger than that of the vectors trained in the target language for the best performance. Thus, we set the dimensionality of English word vectors as 200 and that of Chinese word vectors as 50. 
\par
Table \ref{tb:comparisons} documents the comparisons of the performances between our approach and the competing methods on NLP\&CC 2013 CLSC dataset. From the results, we see that the best performance of co-training is better than that of CL-SCL, but it requires parallel dataset both in training and testing processes. Gui et al. combined co-training and transfer learning strategies. Their method achieved the highest accuracy of 80.1\% in NLP\&CC CLSC task \cite{gui2014cross}. They further improved the accuracy to 80.8\% \cite{gui2015improving} by removing the noise from the transferred samples to avoid negative transfers. Zhou et al. \cite{zhou2015learning} built denoising auto-encoders in two independent views to enhance the robustness to translation errors in the inputs. It integrated the bilingual embedding learning into a unified process, and achieved 80.7\% accuracy. Our approach reaches up to 81.4\% average accuracy with the fixed parameters. In Table \ref{tb:comparisons}, the last row shows the best results of our approach. For Books and Music categories, SCL-OM achieves the best accuracy when $k=120$, $m=300$, while for DVD, it is when $k=110$, $m=200$. The experimental
results show that our approach is competitive with the state-of-the-art in cross-language sentiment classification. 
\subsection{Sensitivity Analysis}
In this section, we analyze the sensitivity of the two important parameters while keeping the others fixed: the number of pivots $m$  and the dimensionality of the cross-lingual representation $k$.
\par
\subsubsection{Number of Pivots $m$:} Figure \ref{fig:m} shows the influence of the number of pivots $m$ on the performance of SCL-OM. The plots show that a small number of pivots can capture a significant amount of the correspondence between $S$ and $T$.
\par
\subsubsection{Dimensionality of the Cross-Lingual Representation $k$:} Figure 3 shows the influence of the dimensionality of the cross-lingual representation $k$ on the performance of SCL-OM. we evaluate SCL-OM when parameter  $k$ varies from 50 to 300. As shown in Figure \ref{fig:k}, the average accuracies generally move upward as $k$ increases. When $k \in (100,150)$, the accuracy reaches the peak value in all three categories, and then the accuracy declines with the increase of $k$.

Furthermore, to gain more insight of the results, we visualize a small part of pivots learned by SCL-OM shown in Table \ref{tb:pivots}. In table \ref{tb:pivots}, the first and the third columns are some examples of pivot features in source language $S$, and the Chinese characters in column two and column four are their corresponding mappings in target language $T$ obtained by our approach. The English word after ``/'' is the translation for non-chinese readers. From this table, we can see that the one-to-many mappings based on the distributed representation of words is more reasonable than the one-to-one mapping by machine translation.
\section{Conclusion}
In this paper, we propose a novel structural correspondence learning method  for cross-lingual sentiment classification with one-to-many mappings. This method employs distributed representation of words to build one-to-many mappings between the pivot features in source language and those in target language. It does not rely on the parallel corpora. This method is evaluated on the NLP\&CC 2013 cross-lingual sentiment analysis dataset, employing English as source language, and Chinese as target language. The experimental results show that our approach is competitive with the state-of-the-art methods in cross-lingual sentiment classification. However, our approach ignores polysemy in the one-to-many mappings. In the future, we will explore the method for learning sense-specific word embedding.
\section{Acknowledgments}
This work is supported in part by Tianjin National Natural Science Foundation for Young Scholars (13JCQNJC00200).

\bibliography{reference}

\begin{thebibliography}{}

\bibitem[\protect\citeauthoryear{Banea \bgroup et al\mbox.\egroup
  }{2008}]{banea2008multilingual}
Banea, C.; Mihalcea, R.; Wiebe, J.; and Hassan, S.
\newblock 2008.
\newblock Multilingual subjectivity analysis using machine translation.
\newblock In {\em Proceedings of the Conference on Empirical Methods in Natural
  Language Processing},  127--135.
\newblock Association for Computational Linguistics.

\bibitem[\protect\citeauthoryear{Blitzer \bgroup et al\mbox.\egroup
  }{2007}]{blitzer2007biographies}
Blitzer, J.; Dredze, M.; Pereira, F.; et~al.
\newblock 2007.
\newblock Biographies, bollywood, boom-boxes and blenders: Domain adaptation
  for sentiment classification.
\newblock In {\em ACL}, volume~7,  440--447.

\bibitem[\protect\citeauthoryear{Blitzer, McDonald, and
  Pereira}{2006}]{blitzer2006domain}
Blitzer, J.; McDonald, R.; and Pereira, F.
\newblock 2006.
\newblock Domain adaptation with structural correspondence learning.
\newblock In {\em Proceedings of the 2006 conference on empirical methods in
  natural language processing},  120--128.
\newblock Association for Computational Linguistics.

\bibitem[\protect\citeauthoryear{Gui \bgroup et al\mbox.\egroup
  }{2013}]{gui2013mixed}
Gui, L.; Xu, R.; Xu, J.; Yuan, L.; Yao, Y.; Zhou, J.; Qiu, Q.; Wang, S.; Wong,
  K.-F.; and Cheung, R.
\newblock 2013.
\newblock A mixed model for cross lingual opinion analysis.
\newblock In {\em Natural Language Processing and Chinese Computing}. Springer.
\newblock  93--104.

\bibitem[\protect\citeauthoryear{Gui \bgroup et al\mbox.\egroup
  }{2014}]{gui2014cross}
Gui, L.; Xu, R.; Lu, Q.; Xu, J.; Xu, J.; Liu, B.; and Wang, X.
\newblock 2014.
\newblock Cross-lingual opinion analysis via negative transfer detection.
\newblock In {\em ACL (2)},  860--865.

\bibitem[\protect\citeauthoryear{Gui \bgroup et al\mbox.\egroup
  }{2015}]{gui2015improving}
Gui, L.; Lu, Q.; Xu, R.; Wei, Q.; and Cao, Y.
\newblock 2015.
\newblock Improving transfer learning in cross lingual opinion analysis through
  negative transfer detection.
\newblock In {\em International Conference on Knowledge Science, Engineering
  and Management},  394--406.
\newblock Springer.

\bibitem[\protect\citeauthoryear{Hu and Liu}{2004}]{hu2004mining}
Hu, M., and Liu, B.
\newblock 2004.
\newblock Mining and summarizing customer reviews.
\newblock In {\em Proceedings of the tenth ACM SIGKDD international conference
  on Knowledge discovery and data mining},  168--177.
\newblock ACM.

\bibitem[\protect\citeauthoryear{Li \bgroup et al\mbox.\egroup
  }{2011}]{li2011semi}
Li, S.; Wang, Z.; Zhou, G.; and Lee, S. Y.~M.
\newblock 2011.
\newblock Semi-supervised learning for imbalanced sentiment classification.
\newblock In {\em IJCAI Proceedings-International Joint Conference on
  Artificial Intelligence}, volume~22,  1826.

\bibitem[\protect\citeauthoryear{Liu}{2012}]{liu2012sentiment}
Liu, B.
\newblock 2012.
\newblock Sentiment analysis and opinion mining.
\newblock {\em Synthesis lectures on human language technologies} 5(1):1--167.

\bibitem[\protect\citeauthoryear{Meng \bgroup et al\mbox.\egroup
  }{2012}]{meng2012cross}
Meng, X.; Wei, F.; Liu, X.; Zhou, M.; Xu, G.; and Wang, H.
\newblock 2012.
\newblock Cross-lingual mixture model for sentiment classification.
\newblock In {\em Proceedings of the 50th Annual Meeting of the Association for
  Computational Linguistics: Long Papers-Volume 1},  572--581.
\newblock Association for Computational Linguistics.

\bibitem[\protect\citeauthoryear{Mikolov, Le, and
  Sutskever}{2013}]{mikolov2013exploiting}
Mikolov, T.; Le, Q.~V.; and Sutskever, I.
\newblock 2013.
\newblock Exploiting similarities among languages for machine translation.
\newblock {\em arXiv preprint arXiv:1309.4168}.

\bibitem[\protect\citeauthoryear{Mogadala and Rettinger}{}]{mogadalabilingual}
Mogadala, A., and Rettinger, A.
\newblock Bilingual word embeddings from parallel and non-parallel corpora for
  cross-language text classification.

\bibitem[\protect\citeauthoryear{Mullen and
  Collier}{2004}]{mullen2004sentiment}
Mullen, T., and Collier, N.
\newblock 2004.
\newblock Sentiment analysis using support vector machines with diverse
  information sources.
\newblock In {\em EMNLP}, volume~4,  412--418.

\bibitem[\protect\citeauthoryear{Pan \bgroup et al\mbox.\egroup
  }{2011}]{pan2011cross}
Pan, J.; Xue, G.-R.; Yu, Y.; and Wang, Y.
\newblock 2011.
\newblock Cross-lingual sentiment classification via bi-view non-negative
  matrix tri-factorization.
\newblock In {\em Pacific-Asia Conference on Knowledge Discovery and Data
  Mining},  289--300.
\newblock Springer.

\bibitem[\protect\citeauthoryear{Pang, Lee, and
  Vaithyanathan}{2002}]{pang2002thumbs}
Pang, B.; Lee, L.; and Vaithyanathan, S.
\newblock 2002.
\newblock Thumbs up?: sentiment classification using machine learning
  techniques.
\newblock In {\em Proceedings of the ACL-02 conference on Empirical methods in
  natural language processing-Volume 10},  79--86.
\newblock Association for Computational Linguistics.

\bibitem[\protect\citeauthoryear{Prettenhofer and
  Stein}{2010}]{prettenhofer2010cross}
Prettenhofer, P., and Stein, B.
\newblock 2010.
\newblock Cross-language text classification using structural correspondence
  learning.
\newblock In {\em Proceedings of the 48th Annual Meeting of the Association for
  Computational Linguistics},  1118--1127.
\newblock Association for Computational Linguistics.

\bibitem[\protect\citeauthoryear{Prettenhofer and
  Stein}{2011}]{prettenhofer2011cross}
Prettenhofer, P., and Stein, B.
\newblock 2011.
\newblock Cross-lingual adaptation using structural correspondence learning.
\newblock {\em ACM Transactions on Intelligent Systems and Technology (TIST)}
  3(1):13.

\bibitem[\protect\citeauthoryear{Turney}{2002}]{turney2002thumbs}
Turney, P.~D.
\newblock 2002.
\newblock Thumbs up or thumbs down?: semantic orientation applied to
  unsupervised classification of reviews.
\newblock In {\em Proceedings of the 40th annual meeting on association for
  computational linguistics},  417--424.
\newblock Association for Computational Linguistics.

\bibitem[\protect\citeauthoryear{Wan}{2009}]{wan2009co}
Wan, X.
\newblock 2009.
\newblock Co-training for cross-lingual sentiment classification.
\newblock In {\em Proceedings of the Joint Conference of the 47th Annual
  Meeting of the ACL and the 4th International Joint Conference on Natural
  Language Processing of the AFNLP: Volume 1-Volume 1},  235--243.
\newblock Association for Computational Linguistics.

\bibitem[\protect\citeauthoryear{Zhai and Zhang}{2016}]{zhai2015semisupervised}
Zhai, S., and Zhang, Z.
\newblock 2016.
\newblock Semisupervised autoencoder for sentiment analysis.
\newblock {\em AAAI}.

\bibitem[\protect\citeauthoryear{Zhang, Chao, and Wang}{2015}]{zhang2015semi}
Zhang, H.; Chao, W.; and Wang, D.
\newblock 2015.
\newblock Semi-supervised learning on cross-lingual sentiment analysis with
  space transfer.
\newblock In {\em Big Data Computing Service and Applications (BigDataService),
  2015 IEEE First International Conference on},  371--377.
\newblock IEEE.

\bibitem[\protect\citeauthoryear{Zhou \bgroup et al\mbox.\egroup
  }{2015}]{zhou2015learning}
Zhou, H.; Chen, L.; Shi, F.; and Huang, D.
\newblock 2015.
\newblock Learning bilingual sentiment word embeddings for cross-language
  sentiment classification.
\newblock ACL.

\bibitem[\protect\citeauthoryear{Zhou, Wan, and Xiao}{2016}]{zhoucross}
Zhou, X.; Wan, X.; and Xiao, J.
\newblock 2016.
\newblock Cross-lingual sentiment classification with bilingual document
  representation learning.

\end{thebibliography}
\bibliographystyle{aaai}
\end{document}